\documentclass{article}
\usepackage{amsmath}
\usepackage{natbib}

\begin{document}

\title{How to show a probabilistic model is better}
\date{}
\author{Mithun Chakraborty, Sanmay Das, Allen Lavoie\\
Washington University in St.~Louis\\
\{mithunchakraborty,sanmay,allenlavoie\}@wustl.edu}
\maketitle

\begin{abstract}
% Meant more as a tutorial
  We present a simple theoretical framework, and corresponding
  practical procedures, for comparing probabilistic models on real
  data in a traditional machine learning setting. This framework is
  based on the theory of proper scoring rules, but requires only basic
  algebra and probability theory to understand and verify. The
  theoretical concepts presented are well-studied, primarily in the
  statistics literature. The goal of this paper is to advocate their
  wider adoption for performance evaluation in empirical machine
  learning.
\end{abstract}
\section{Why probabilistic predictions?}
When a model is applied to a situation where uncertainty is inherent
(e.g.~predicting a biased coin flip, or a user's next click), a
probability distribution should be its output. Accurate probability
distributions provide more information than point predictions, and are
the natural product of Bayesian models. Our goal is not to advocate
probabilistic models \emph{per se}, but to show in an accessible way
that their output can be evaluated rigorously with no more difficulty
than deterministic labelings in classification problems.
\section{Comparing models}
Where do observations come from? They are based on the state of the
world. This state describes the situation in which a model is asked to
make a prediction.
\begin{equation}
  \sigma \sim S
  \label{eqn:drawstate}
\end{equation}
The support of this distribution $S$ over states is likely infinite
and uncountable. If we are predicting the weather then a state
$\sigma$ includes a description of physical phenomena that could
affect future weather patterns. If we are predicting which ad a user
will click on, a state includes factors influencing the user's
decision: personality, past history, web page design, and so on. The
distribution is entirely theoretical, and need never be described
formally.

Based on the state of the world $\sigma$, an outcome is observed and
recorded. However, the outcome is not necessarily implied
deterministically by $\sigma$. Rather, there is a distribution over
possible outcomes:
\begin{equation}
  x \sim f_\sigma
  \label{eqn:drawobs}
\end{equation}
This includes the possibility of a degenerate distribution
(probability 1 on a single outcome), but does not require
it. Uncertainty could stem from true randomness (e.g.~quantum noise)
or from ignorance (e.g.~the model does not know what the user ate for
breakfast). The noise distribution $f_\sigma$ is again entirely
theoretical, and need never be described.

Equations~\eqref{eqn:drawstate} and \eqref{eqn:drawobs} define a
generative framework for observations. When scoring the probabilistic
predictions of a model, we will typically have a single observation
from each of many \emph{different} states of the world $\sigma_1
\ldots \sigma_n$ (although states drawn multiple times pose no
problem). That is, we have a set $X$ of $n$ observations:
\begin{equation}
  X = \{x_{\sigma_1} \ldots x_{\sigma_n}\}
\end{equation}
For convenience, we will assume that these observations are discrete,
but a generalization to real-valued observations is
possible. Corresponding to each of these observations are predictions
from each of the models we are evaluating. For simplicity, we assume
two models, $g$ and $k$.
\begin{align}
  G = \{g_{\sigma_1} \ldots g_{\sigma_n}\}\\
  K = \{k_{\sigma_1} \ldots k_{\sigma_n}\}
\end{align}
Here $g_{\sigma_1}$ is the distribution that model $g$ predicts in the
state $\sigma_1$ where the observed outcome is $x_{\sigma_1}$, and
likewise for the remainder of the observations and for model $k$. The
theoretical assumption is that $x_{\sigma_i} \sim f_{\sigma_i}$, but
the states $\sigma_i$ need no description for the purposes of model
evaluation and we never need to construct $f_{\sigma_i}$
explicitly. To say that model $g$ is ``better'' than model $k$, we
would like to conclude that it has a lower divergence from the true
distribution $f$ in expectation for some divergence function $d$:
\begin{equation}
  E_{\sigma \sim S}[d(f_\sigma || g_\sigma)] < E_{\sigma \sim S}[d(f_\sigma || h_\sigma)]
  \label{eqn:objective}
\end{equation}
Since we have a finite number $n$ of samples, we can only determine
probabilistically if this inequality holds. Examples of $d$ for which
this estimation task is possible using only $X$, $G$, and $K$ are
squared Euclidean distance $d(p || q) = ||p - q||^2$ and KL-divergence
$d(p || q) = \sum_j p_j \ln\frac{p_j}{q_j}$. However, it is not
immediately obvious how the truth of the inequality in
\eqref{eqn:objective} can be evaluated, even probabilistically,
\emph{without access to the true distributions $f_{\sigma_1}, \ldots,
  f_{\sigma_n}$}. However, only simple algebra is required. For
KL-divergence, we first approximate the expectation of the log
probability assigned by model $g$ (the derivation for $k$ is
identical) to the true outcome $x$, that is:
\begin{equation}
  E_{\sigma \sim S}\left [E_{x \sim f_{\sigma}}[-\ln(g_{\sigma, x})] \right]
  \label{eqn:scoringrule}
\end{equation}
This expression can be approximated from $G$ and $X$:
\begin{equation}
  -\frac{1}{n}\sum_{i=1}^n \ln(g_{\sigma_i, x_{\sigma_i}})
  \label{eqn:empirical}
\end{equation}
Where $g_{\sigma_i, x_{\sigma_i}}$ is the probability that model $g$
assigned to the true outcome $x_{\sigma_i}$ (corresponding in the
theoretical model to state $\sigma_i$). The trick is that
\eqref{eqn:scoringrule} is equivalent to expected KL-divergence plus a
constant:
\begin{align*}
  &=E_{\sigma \sim S}\left [\sum_j f_{\sigma,j} (-\ln g_{\sigma, j}) \right]\\
  &=E_{\sigma \sim S}\left [\sum_j f_{\sigma,j} (-\ln g_{\sigma, j} + \ln f_{\sigma,j} - \ln f_{\sigma,j}) \right]\\
  &=E_{\sigma \sim S}\left [\sum_j \left (f_{\sigma,j} \ln\frac{f_{\sigma,j}}{g_{\sigma, j}} - f_{\sigma,j} \ln f_{\sigma,j}\right) \right]\\
  &=E_{\sigma \sim S}\left [d_{\mathrm{KL}}(f_\sigma || g_\sigma) + H(f_\sigma) \right]
\end{align*}
Here $H(f) = -\sum_j f_j \ln f_j$ is the Shannon entropy of $f$. Since
$H(f_\sigma)$ is independent of a model's predictions, differences in
\eqref{eqn:scoringrule} between models $g$ and $k$ must be due to
differences in the expected KL-divergences $E_{\sigma \sim
  S}[d_{\mathrm{KL}}(f_\sigma || g_\sigma)]$ and $E_{\sigma \sim
  S}[d_{\mathrm{KL}}(f_\sigma || k_\sigma)]$. The only remaining
complication is the finite sample: how can we be sure that observed
differences in \eqref{eqn:empirical} are due to differences in
\eqref{eqn:scoringrule}?

This is a standard statistical task: we have a set of $n$ paired
samples $(-\ln g_{\sigma_i,x_{\sigma_i}}, -\ln
k_{\sigma_i,x_{\sigma_i}})$ related by the state of the world
$\sigma_i$ for each sample, and want to test whether the expectation
of the $g$ samples is significantly less than that of the $k$ samples
(meaning $g$ is a better model). A paired t-test or the Wilcoxon
signed-rank test (although it tests the median rather than the mean) are
reasonable options.

This simple algebraic trick comes out of the theory of proper scoring
rules (see \citet{gneiting2007strictly} for a thorough
survey). Scoring rules were developed to incentivize true reporting of
probabilities by experts: first a report is solicited in the form of a
probability distribution $q$, then an outcome is observed. The expert
is paid based on their report and the outcome, according to the
scoring rule. A proper scoring rule incentivizes an expert to report
truthfully (which is not the case if the expert is paid e.g.~$q_i$ for
an outcome $i$, often referred to as the naive scoring rule). Any
proper scoring rule has an associated divergence function, which for
the logarithmic scoring rule ($\ln q_i$ for an observed outcome $i$)
is KL-divergence. The divergence function associated with the
quadratic scoring rule $2 q_i - ||q||^2$ is squared Euclidean
distance, which can also be derived with only simple algebra:
\begin{align}
  &E_{\sigma \sim S}\left [E_{x \sim f_{\sigma}} \left [-2 g_{\sigma, x} + ||g_\sigma||^2 \right ] \right]\label{eqn:quadscore}\\
  =&E_{\sigma \sim S}\left [-2 f_{\sigma} \cdot g_{\sigma, j} + ||g_\sigma||^2 + ||f_{\sigma}||^2 - ||f_{\sigma}||^2\right]\\
  =&E_{\sigma \sim S}\left [||f_{\sigma} - g_\sigma||^2 - ||f_{\sigma}||^2\right]\\
\end{align}
As with the logarithmic scoring rule, we get a divergence function
$||f_{\sigma} - g_{\sigma}||^2$ and a generalized entropy term
$||f_{\sigma}||^2$ which again is independent of the model's
predictions.
\section{Procedure summary}
While theoretically justifying probabilistic model comparisons is
slightly tedious, the procedure could not be simpler. To summarize:
\begin{itemize}
\item
  For every held-out observation, score each model's predicted
  distribution $q$: $-\ln(q_i)$ for logarithmic, or $-2 q_i + ||q||^2$
  for quadratic given that outcome $i$ is observed
\item
  Perform a (typically paired) statistical test to determine whether
  the scores for one model are significantly lower than those for the
  other, lower indicating a better model
\end{itemize}
When comparing more than two models, perform as many pairwise tests as
necessary. The ``figure of merit'' for a model is its mean score
(e.g.~\eqref{eqn:empirical} for the logarithmic scoring rule), lower
implying less divergence from the unobserved true distributions of
observations and therefore a better model (modulo noise in the
estimate).
\section{Choosing a divergence function}
We have presented two of the most common choices for scoring rules,
quadratic and logarithmic, corresponding to evaluation with squared
Euclidean distance and KL-divergence respectively. The former is of
interest when KL-divergence is too quick to dismiss models which put
zero probability on observed outcomes. While these models are clearly
``wrong,'' i.e.~provably not reporting the true distribution of
observations, this is usually not our main concern (``all models are
wrong, some are useful''). Often we want to compare against model-free
baselines (e.g.~observed frequencies) which do report zero probability
on observed outcomes, and adding parameters to hedge their reports is
undesirable. When this is not an issue, KL-divergence is desirable due
to its popularity and connections to information theory.

Other proper scoring rules exist, and their divergence functions can
be used in model comparisons just as for the logarithmic and quadratic
scoring rules. For example, cosine similarity corresponds to the
spherical scoring rule. In general, any Bregman divergence is a
feasible way of comparing probabilistic models (i.e.~has an associated
proper scoring rule).
\section{Alternatives}
One popular method of scoring probabilistic models is
\emph{perplexity}, which is simply an exponentiated version of
\eqref{eqn:empirical}. This exponentiation rewards slight
over-reporting of high probability events, but the effect diminishes
rapidly with increasing dataset size. Nonetheless, it is theoretically
preferable to use the un-exponentiated version for model comparisons.

There are many popular ways of scoring non-probabilistic predictions
based on classification accuracy, precision and recall, and so
on. These methods can be applied to probabilistic models, for example
by ranking outcomes by their reported probability. However, such
procedures discard much of the information probabilistic predictions
provide, and so are generally less desirable when choosing between
probabilistic models.

\bibliographystyle{unsrtnat}
\bibliography{modelscoring}

\end{document}